\documentclass[letterpaper, 10pt, conference]{ieeeconf}  
\IEEEoverridecommandlockouts  
\usepackage{amsmath,amssymb,amsfonts,amsthm}

\usepackage{algorithm}
\usepackage{algpseudocode}
\makeatletter
\let\orig@fs@ruled\fs@ruled
\def\fs@ruled{%
  \orig@fs@ruled
  \let\alg@orig@fs@pre\@fs@pre
  \def\@fs@pre{\vspace*{2mm}\alg@orig@fs@pre}%
}
\makeatother

\usepackage{graphicx}            
\usepackage{xcolor}              
\usepackage{threeparttable}      
\usepackage{booktabs}
\usepackage{array}               
\newcolumntype{C}[1]{>{\centering\arraybackslash}p{#1}}  

\usepackage{cite}                
\usepackage{hyperref}            

\usepackage{comment}             

\theoremstyle{remark}

\newcommand{\figref}[1]{\mbox{Fig.~\ref{#1}}}
\newcommand{\tabref}[1]{\mbox{Table~\ref{#1}}}
\newcommand{\secref}[1]{\mbox{Section~\ref{#1}}}

\newcommand{\titledparagraph}[1]{\vspace{1mm}\noindent\textbf{#1}}

\newif\ifshowflags
\showflagstrue

\title{\LARGE \bf
Perceptive Variable-Timing Footstep Planning for Humanoid Locomotion on Disconnected Footholds
}
\author{Zhaoyang Xiang, Upama Pant, and Ayonga Hereid
\thanks{Mechanical and Aerospace Engineering, The Ohio State University, Columbus, OH 43210, USA. {\tt\footnotesize (xiang.295, pant.46, hereid.1)@osu.edu.}}
}

\begin{document}

\maketitle

\begin{abstract}
Walking over disconnected footholds requires selecting landing regions while adapting step timing to the robot's motion. We present a perceptive model predictive planner that combines discrete region selection, foot placement, and current-step duration in a mixed-integer quadratic program using step-to-step Divergent Component of Motion (DCM) dynamics. Onboard depth data are fused into a probabilistic heightmap in the stance foot frame. Convex steppable regions are extracted and shrunk by margins that increase with estimated height uncertainty. Linear constraints impose capturability-based bounds on DCM evolution and limit swing-foot retargeting using the remaining commanded swing time. Within-step replanning updates the initial DCM from measurements to respond to model mismatch and disturbances. We evaluate the framework on Digit in simulation over randomized footholds and on hardware over separated plates, with mean planner solve times of 11.4~ms and 4.0~ms, respectively.
\end{abstract}

\section{Introduction}
Bipedal locomotion over gaps, obstacles, and other unsteppable areas often requires walking on a disconnected set of footholds. The controller must select a landing region and adapt step timing to the robot's motion. Reduced-order template models~\cite{englsberger2015threedimensional,gong2022zero} support model predictive control (MPC) of Center of Mass (CoM) motion and footsteps in real time~\cite{wieber2006trajectory}. On disconnected footholds, region selection is discrete, while step duration determines the growth of the unstable component of motion before touchdown. Without a prior terrain map, admissible footholds must also be extracted online from onboard perception.

Footstep control on specified footholds is well studied. Control barrier functions and gait libraries guide the swing foot to a prescribed landing point~\cite{nguyen2017dynamic,dai2022bipedal}, while mixed-integer programs select a region from a union of convex sets and optimize placement within it~\cite{deits2014footstep,aceituno-cabezas2018simultaneous,fey20243d}. These formulations take landing targets or admissible regions as inputs.
Step timing provides another way to adapt motion to restricted footholds. Switching-time optimization in whole-body MPC~\cite{katayama2022wholebody}, adaptive-frequency walking on uneven stepping stones~\cite{li2023dynamic}, and a DCM-based controller that adapts step duration for accurate foot placement~\cite{xiang2025adaptive} incorporate timing adaptation into locomotion control. Li et al.~\cite{li2025gaitnet} use a learned gait-frequency network within sequential MPC to update step duration along with foot placement and contact forces. However, their evaluation assumes known terrain and exposes only a one-step preview to the controller.

\begin{figure}[t]
    \vspace{2mm}
    \centering
    \includegraphics[width=\linewidth]{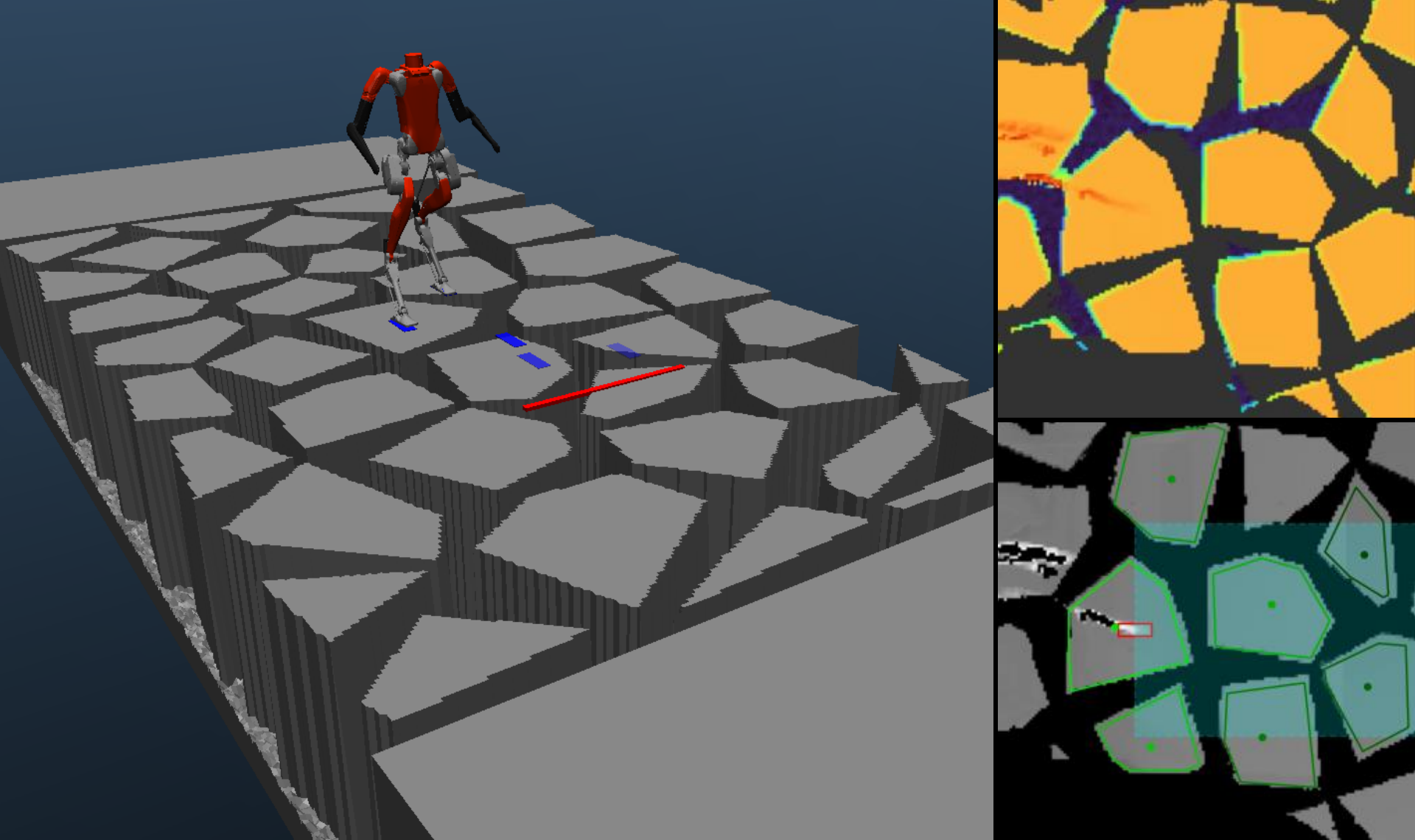}
    \caption{(Left) Snapshot of one of the random simulation terrains for all ablation tests. The blue rectangles show the planned step position for the previewed steps, and the red bar shows the goal position in the sagittal direction. (Top Right) Local heightmap colored by height. (Bottom Right) Steppable cells (gray), selection window (shaded rectangle), stance foot (red), and convex regions passed to the planner (green polygons).}
    \label{fig:sim_snapshot}
    \vspace{-2mm}
\end{figure}

Perceptive locomotion obtains foothold constraints from onboard depth cameras or LiDAR. Probabilistic elevation maps fuse depth measurements into a local height belief~\cite{fankhauser2018probabilistic,miki2022elevation}, from which planar or polygonal regions can be extracted~\cite{mishra2024efficient}. Grandia et al.~\cite{grandia2023perceptive} combine terrain segmentation with MPC and shrink region boundaries to provide safety margins. A further design choice is how estimated height uncertainty should affect the admissible landing area.
Acosta and Posa~\cite{acosta2023bipedal,acosta2025perceptive} combine online terrain perception with mixed-integer footstep control for underactuated bipeds. Their approach jointly optimizes discrete foothold selection, placement, ankle torque, and the remaining duration of the current stance in one mixed-integer quadratic program (MIQP). They linearize the Angular Momentum Linear Inverted Pendulum (ALIP) solution with respect to remaining stance time. In contrast, our DCM formulation uses an exponential timing variable, so the current-step prediction is exact for the template model and affine in the decisions, without linearization. Moreover, we combine this formulation with directional capturability bounds and uncertainty-dependent region margins.

\begin{figure*}[!t]
    \vspace{2mm}
    \centering
    \includegraphics[width=\textwidth]{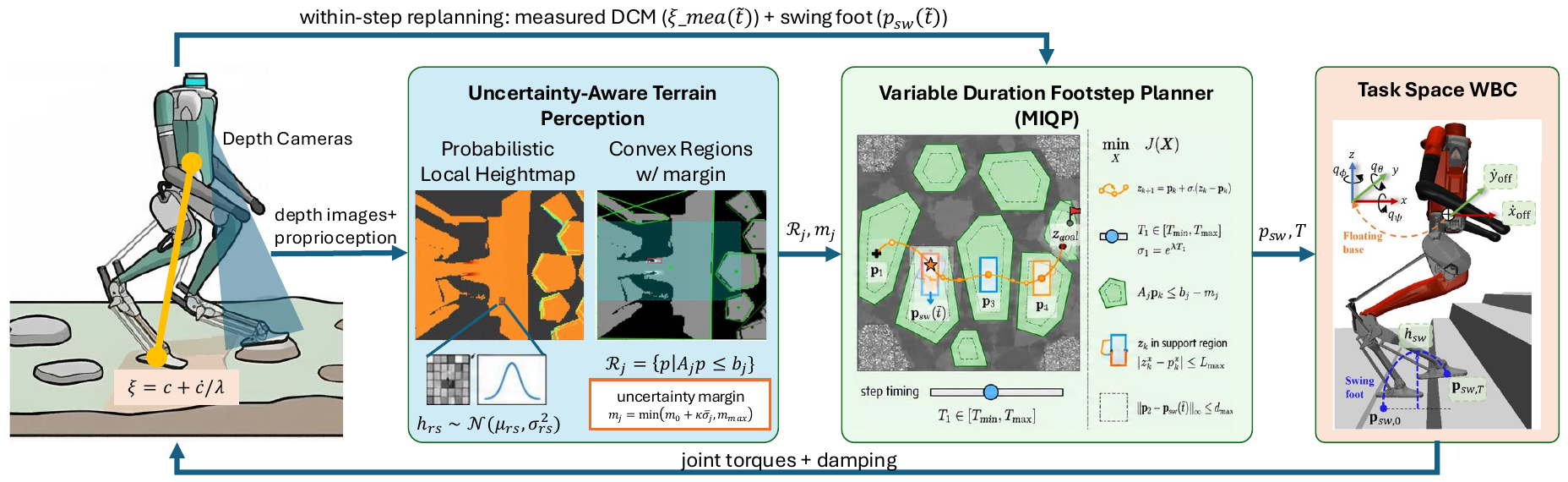}
    \caption{Overview of the framework. Perception builds a probabilistic local heightmap from onboard depth cameras and extracts convex steppable regions. Regions with less certain height are shrunk more. The MIQP planner chooses the next footholds and the duration of the current step. The whole-body controller (WBC) tracks a swing-foot trajectory to the planned foothold and computes joint torques. During each swing, the planner is solved again from the measured DCM and swing-foot position.}
    \label{fig:overview}
    \vspace{-2mm}
\end{figure*}

Learning-based controllers also enable dynamic walking on sparse footholds and parkour courses. Duan et al.~\cite{duan2022learning} combine learned footstep tracking and prediction of achievable steps with overhead vision. Other policies map onboard depth images or elevation maps to joint commands and learn where to step from terrain rewards~\cite{zhuang2024humanoid,wang2025beamdojo,liu2026worldmodel}. Whole-body parkour policies track human motion data and use onboard depth to adapt these motions to obstacles~\cite{zhuang2026deep,wu2026perceptive}. Reduced-order models can also generate reference steps that guide training~\cite{crismariu2026march,dai2026planc}. However, a trained policy selects footholds and step timing implicitly. Landing inside a perceived foothold and keeping balance are encouraged during training, not explicitly imposed at run time.

We present a perceptive DCM-based planner that integrates footstep selection and current-step timing with locally perceived foothold constraints. The framework:
(i) jointly optimizes region membership, foot placement, and current-step duration in an MIQP;
(ii) bounds lateral and sagittal DCM evolution using capturability-based constraints;
(iii) extracts convex steppable regions from onboard depth data in the stance foot frame, so it does not depend on a world-frame pose estimate;
(iv) accounts for perception uncertainty by keeping footsteps farther from the edges of regions whose estimated height is less certain; and
(v) replans during each swing using measured motion as feedback, while limiting how far the landing target can move given the swing foot's position and the time left before touchdown.
In simulation, Digit crosses a randomized stepping-stone field with irregular gaps in both horizontal directions, at a forward speed of about 1~m/s over much of the crossing. Without timing adaptation, with a shorter preview, or without the capturability bounds, the robot fails to cross in the reported runs. On hardware, Digit uses body-mounted depth cameras to cross separated plates 0.18~m high with gaps averaging 0.3~m, including arrangements with one plate rotated by $\pm15^\circ$. The planner solves in 4.0~ms on average.

The rest of the paper is organized as follows. \secref{sec:perception} describes the depth-based perception that produces the convex steppable regions. \secref{sec:planning} formulates the MIQP over those regions. \secref{sec:results} reports simulation and hardware results on Digit, and \secref{sec:conclusion} concludes the paper.

\section{Probabilistic Ego-centric Terrain Perception} \label{sec:perception}
The planner needs steppable regions and an estimate of their uncertainty. We obtain both from onboard depth data by maintaining a local probabilistic heightmap and extracting convex regions from it. \figref{fig:overview} shows where these regions enter the planner and the whole-body controller.

\subsection{Probabilistic Local Heightmap}
\titledparagraph{Map Representation and Frame Update.}
At each perception cycle, $N_c$ body-mounted depth cameras cover the forward and backward directions. Their depth data are back-projected into 3D point clouds in the respective camera frames. Using the estimated floating base pose, these points are transformed to the stance foot frame as $\mathcal{PC}_s$.

A fixed-size 2.5D elevation grid $\mathcal{H}$ with resolution $\Delta r$ is maintained in this frame, with one height distribution per horizontal cell. Each cell $\mathcal{H}(r,s)$, indexed by row $r$ and column $s$, stores a Gaussian height belief $h_{rs}\sim\mathcal{N}(\mu_{rs},\sigma^2_{rs})$, where $\mu_{rs}$ is the estimated terrain height and $\sigma^2_{rs}$ represents uncertainty in that height. It also stores an observation count $n_{rs}$ and a transform count $\eta_{rs}$ that tracks frame-motion resampling operations. Each cycle compensates frame motion when a previous frame exists, clears stale cells, fuses current observations, and increases uncertainty in unobserved cells.

To retain previous observations as the stance frame moves, we transform the existing grid into the new frame. The stance frame changes only at touchdown, so within a step the transform is the identity and the map is left unchanged. At each touchdown the whole-body controller latches the origin and yaw of the departing stance frame expressed in the new one, evaluated from leg kinematics at that instant, and passes them to perception. Let $\Delta\mathbf{R}^z$ and $\Delta\mathbf{t}$ denote this latched rotation and translation. A cell center moves according to
\begin{align} \label{eq:frame_transform}
    \mathbf{p}_{\mathrm{new}} = \Delta\mathbf{R}^z\, \mathbf{p}_{\mathrm{old}} + \Delta\mathbf{t}.
\end{align}
The update requires no world-frame pose and no point-cloud registration. Because the latch is a single kinematic relation evaluated at one instant, rather than a difference between two pose estimates separated in time, its horizontal and yaw components do not accumulate error across steps. When several old cells map to the same new cell, the cell with the greatest height is retained.

To avoid unnecessary resampling and reduce confidence in repeatedly propagated data, the map update applies three safeguards before measurement fusion:
\begin{enumerate}
    \item \emph{Motion threshold.} Resampling is skipped when the latched step's translation and yaw fall below prescribed thresholds, avoiding unnecessary interpolation for a near-in-place step.
    \item \emph{Resampling penalties.} After each grid transformation, cell variance is multiplied by $\alpha_\sigma>1$ and observation count by $\alpha_n<1$.
    \item \emph{Staleness clearing.} Cells whose transform count or age exceeds their respective limits are cleared entirely.
\end{enumerate}

\titledparagraph{Observation Variance Model.}
The observation model assigns greater uncertainty to distant points and grazing views. For a point $\mathbf{p}_{pc}$ in the camera frame at distance $d=\|\mathbf{p}_{pc}\|$, the observation variance is
\begin{align} \label{eq:variance_model}
    \sigma^2_{\mathrm{obs}} = \sigma_0^2 \, \frac{d^2}{\max\!\bigl(\cos\theta,\, \epsilon\bigr)},
\end{align}
where $\sigma_0$ is the baseline noise at unit distance and perpendicular viewing, $\cos\theta=|p^z_{pc}|/d$ represents a locally horizontal surface assumption, and $\epsilon$ prevents singularities at grazing incidence.

\titledparagraph{Bayesian Cell Update.}
When $Q>1$ depth points land in the same cell within a single frame, their heights are first aggregated by inverse-variance weighting. Let $z_q$ be the height of point $q$ in the stance foot frame and $\sigma_q^2$ its observation variance from~\eqref{eq:variance_model}. With weights $w_q=1/\sigma_q^2$, the aggregate height and variance are
\begin{align} \label{eq:preagg}
    \mu_{\mathrm{agg}} = \frac{\sum_{q=1}^{Q} w_q z_q}{\sum_{q=1}^{Q} w_q}, \quad
    \sigma^2_{\mathrm{agg}} = \frac{1}{\sum_{q=1}^{Q} w_q},
\end{align}
so observations with smaller variance receive greater weight. The aggregated observation is then fused with the cell prior through a scalar Kalman update:
\begin{align} \label{eq:cell_update}
    K &= \sigma^2_{rs} \big/ \bigl(\sigma^2_{rs} + \sigma^2_{\mathrm{agg}}\bigr), \\
    \mu_{rs} &\leftarrow \mu_{rs} + K\bigl(\mu_{\mathrm{agg}} - \mu_{rs}\bigr), \\
    \sigma^2_{rs} &\leftarrow (1 - K)\,\sigma^2_{rs}.
\end{align}
The gain $K$ gives more weight to the new observation when its variance is small relative to the prior variance. A variance floor prevents overconfidence. Upon receiving a fresh observation, $\mu_{rs}$ is reset to zero so that the cell is treated as freshly observed when checking staleness.

\titledparagraph{Uncertainty Growth in Unobserved Cells.}
After measurement fusion, cells not observed in the current frame undergo variance growth at a motion-dependent rate:
\begin{align} \label{eq:uncertainty_growth}
    \sigma^2_{rs} \leftarrow \gamma_{\mathrm{eff}} \,\sigma^2_{rs}, \quad
    \gamma_{\mathrm{eff}} = 1 + (\gamma_0 - 1)\left(1 + \frac{\|\Delta\mathbf{t}\|}{d_{\mathrm{ref}}}\right),
\end{align}
where $\gamma_0>1$ is the base variance-growth factor and $\|\Delta\mathbf{t}\|$ is the stance frame displacement from~\eqref{eq:frame_transform}. The reference displacement $d_{\mathrm{ref}}$ normalizes the motion magnitude. Since the stance frame moves only at touchdown, $\|\Delta\mathbf{t}\|=0$ within a step and $\gamma_{\mathrm{eff}}=\gamma_0$, so unobserved cells still lose confidence while the frame is fixed. As inter-frame translation increases, their variance grows faster. This motion-dependent inflation reduces confidence in cells carried through map updates without fresh observations.

\subsection{Convex Region Constraints}
\begin{figure}[t]
    \vspace{2mm}
    \centering
    \includegraphics[width=\linewidth]{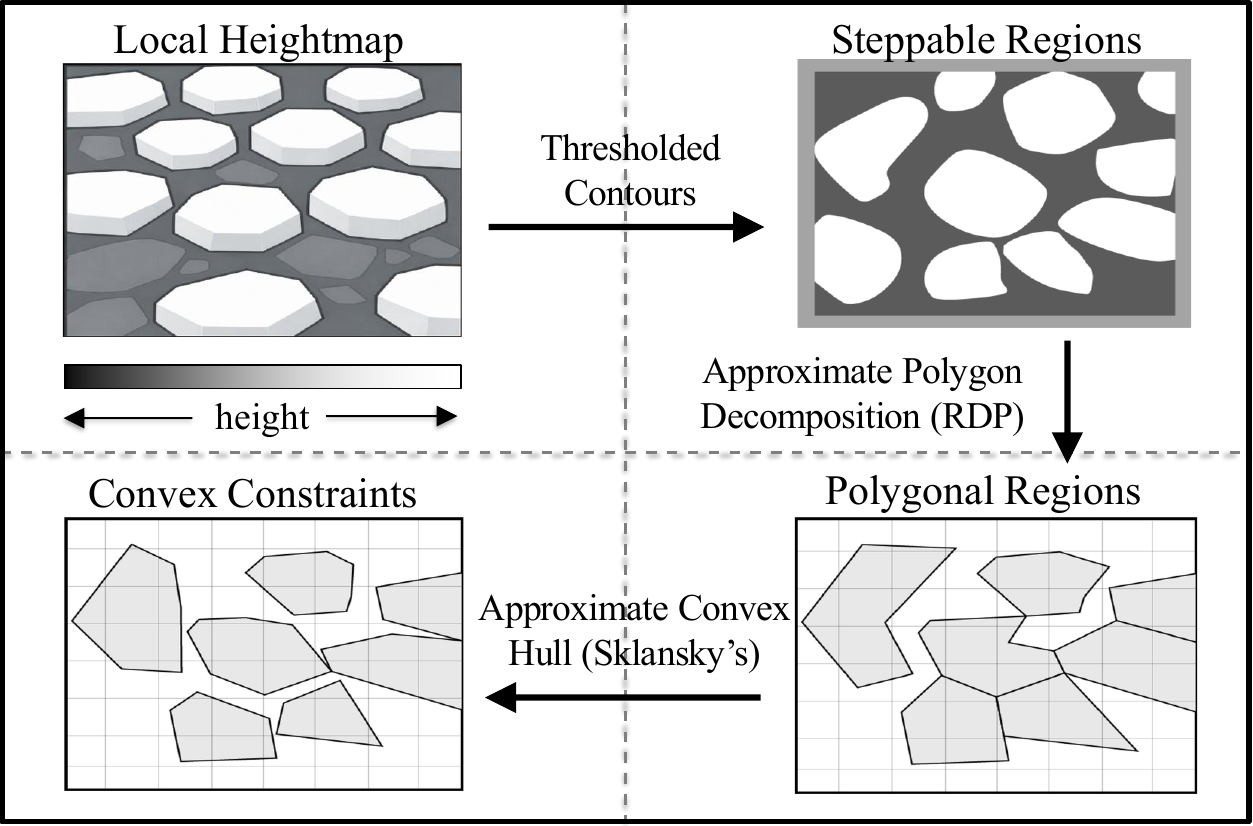}
    \caption{From heightmap to steppable regions. Top left: the local heightmap, shaded by height. Top right: candidate steppable cells after thresholding. Bottom right: contours simplified to polygons by RDP. Bottom left: convex hulls represented as linear region constraints.}
    \label{fig:region_extraction}
    \vspace{-2mm}
\end{figure}

\titledparagraph{Region Extraction.}
To express admissible footholds as linear constraints, we extract convex regions from the heightmap using OpenCV primitives~\cite{bradski2000opencv}, as shown in~\figref{fig:region_extraction}. Thresholding identifies candidate steppable cells, whose boundaries form contours. After removing small contours, the Ramer--Douglas--Peucker algorithm (RDP) simplifies each contour to a polygon with a bounded number of edges. Its tolerance is tied to grid resolution to suppress sensor noise. Sklansky's algorithm~\cite{sklansky1982finding} then forms the convex hull, which may fill concavities in the detected region. Each hull is represented by $\mathcal{R}_j=\{\mathbf{p}\mid\mathbf{A}_j\mathbf{p}\le\mathbf{b}_j\}$. Each row of $\mathbf{A}_j$ is a unit outward normal to one edge, so entries of $\mathbf{b}_j$ are signed distances from the origin.

\titledparagraph{Region Selection and Planner Output.}
To limit the number of regions considered by the planner, we retain at most $M_{\max}$ candidates. When more are detected, a rectangular selection window is aligned with the intended walking direction. One offset, proportional to the commanded goal, shifts it forward or backward; a second extends it backward at low forward speed to include possible step-back locations. Since the stance region is always retained, other regions with sufficient window overlap are sorted by proximity until the capacity is reached. We use Sutherland--Hodgman polygon clipping to determine the overlap with the window.

The cell variances account for observation noise through measurement fusion and are adjusted for resampling and missed observations. For each retained region, $\bar\sigma_j$ is the mean height standard deviation of the heightmap cells inside $\mathcal{R}_j$. Perception supplies the planner with $\{(\mathbf{A}_j,\mathbf{b}_j,\bar\sigma_j)\}_{j=1}^{M}$, where $M\le M_{\max}$. The geometry defines the union of steppable regions. The planner converts the height-uncertainty summary into a heuristic inward margin.

\section{Variable Time Foot Placement via MIQP} \label{sec:planning}
On stepping-stone terrain, a footstep plan must choose both where to land and when to end the current step. We formulate an MIQP that combines DCM predictions with the steppable regions from~\secref{sec:perception}, capturability-based bounds, and swing reachability. The plan is periodically updated during the swing. Each update revises the landing position and step duration, which change the swing-foot reference tracked by the whole-body controller.

\subsection{Prediction Model and Decision Variables}
The DCM describes the unstable component of the horizontal CoM motion~\cite{englsberger2015threedimensional}. For CoM position $\mathbf{c}\in\mathbb{R}^2$ and velocity $\dot{\mathbf{c}}$, it is $\boldsymbol{\xi}=\mathbf{c}+\dot{\mathbf{c}}/\lambda$, where $\lambda=\sqrt{g/z_c}$, $g$ is gravitational acceleration, and $z_c$ is the constant CoM height. Let $\boldsymbol{z}_k\in\mathbb{R}^2$ denote the DCM at the start of step $k$, $\mathbf{p}_k\in\mathbb{R}^2$ the stance foot location used as the center of pressure during that step, and $T_k$ its duration. For an $N$-step preview and timing gain $\sigma_k=e^{\lambda T_k}$, the step-to-step dynamics~\cite{xiang2025adaptive} can be written as
\begin{align} \label{eq:dcm_discrete_dynamics}
 \boldsymbol{z}_{k+1} = \mathbf{p}_k + \sigma_k(\boldsymbol{z}_k-\mathbf{p}_k),
 \quad k=1,\dots,N.
\end{align}
We call $\boldsymbol{z}_k-\mathbf{p}_k$ the DCM offset: the position of the DCM relative to the stance foot at the start of step $k$. Since $\sigma_k>1$, this offset grows by the factor $\sigma_k$ by the end of the step.

\titledparagraph{Timing and Initial Conditions.}
Optimizing every duration would introduce products of timing, state, and foot-placement variables in~\eqref{eq:dcm_discrete_dynamics}. To keep the prediction affine in the decisions, we optimize only the current step's gain $\sigma_1$ and fix later gains to $\sigma_{\mathrm{nom}}=e^{\lambda T_{\mathrm{nom}}}$:
\begin{align} \label{eq:sigma_fix}
 \sigma_k=\sigma_{\mathrm{nom}},\quad k=2,\dots,N.
\end{align}
At each solve, the current stance position $\mathbf{p}_1=\mathbf{0}$ and the initial DCM $\boldsymbol{z}_1$ are fixed inputs in the stance foot frame. The latter is reconstructed from the latest measured DCM during replanning. Thus, the first transition multiplies $\sigma_1$ only by known quantities; later transitions use the fixed gain $\sigma_{\mathrm{nom}}$. Both cases are affine: each transition is linear in the decisions plus known constants. The admissible durations give
\begin{align} \label{eq:sigma_bounds_planning}
 \sigma_{\min}=e^{\lambda T_{\min}}\le\sigma_1
 \le e^{\lambda T_{\max}}=\sigma_{\max},
\end{align}
with $T_{\mathrm{nom}}\in[T_{\min},T_{\max}]$. These limits restrict timing according to actuator and swing-motion requirements, and later steps cannot adjust duration to trade time for reach. Although later durations are fixed, later foot placements and region assignments remain decisions. A candidate landing may fit one stone yet leave no feasible continuation to the next stones. Preview therefore constrains the current timing and foothold through the subsequent steps that the model and terrain constraints allow.

\titledparagraph{Foot Placements and Strides.}
Over the $N$-step horizon, $\mathbf{p}_2,\dots,\mathbf{p}_N$ are free foot placements. The state $\boldsymbol{z}_2$ is the DCM at the transition to this next support. Since $\mathbf{p}_1=\mathbf{0}$, its offset from the landing foot is
\begin{align} \label{eq:touchdown_offset}
 \boldsymbol{z}_2-\mathbf{p}_2=\sigma_1\boldsymbol{z}_1-\mathbf{p}_2.
\end{align}
Changing the duration changes the DCM reached at touchdown; changing the landing position changes its offset from the next support. These decisions must be made together because the available steppable regions restrict where that support can be placed.

We express stride length and width through the auxiliary variables
\begin{align}
 s_k^x &= p_{k+1}^x-p_k^x, \nonumber\\
 s_k^y &= (-1)^{k+\ell}(p_{k+1}^y-p_k^y),
 \quad k=1,\dots,N-1, \label{eq:slacks}
\end{align}
where $\ell=1$ if the current stance foot is the right foot and $\ell=0$ if it is the left foot. Thus $k+\ell$ is even on right-stance steps and odd on left-stance steps. The alternating sign makes positive width mean the same outward foot separation for either stance. The same constraints therefore apply whichever foot is in stance when the planner is called. Strides satisfy
\begin{align} \label{eq:stride_bounds}
 L_{\min}\le s_k^x\le L_{\max},\qquad
 W_{\min}\le s_k^y\le W_{\max}.
\end{align}
The bounds reflect kinematic reach and foot clearance and can be tightened for terrain restrictions. Using strides also lets the objective express a nominal gait independently of absolute foot position. The continuous decision variables are
\begin{align*}
 \mathcal{X}=\big\{&\{\mathbf{p}_k\}_{k=2}^{N},\;
 \{\boldsymbol{z}_k\}_{k=2}^{N+1},\;\sigma_1,\;
 \{s_k^x,s_k^y\}_{k=1}^{N-1}\big\}.
\end{align*}

\subsection{MIQP Formulation}
\titledparagraph{Objective.}
The objective balances progress toward a DCM target with departures from nominal timing and strides. Let $\boldsymbol{z}_{\mathrm{goal}}\in\mathbb{R}^2$ be the prescribed target, set ahead of the robot in the experiments, and let $s_{\mathrm{nom}}^x,s_{\mathrm{nom}}^y$ be the nominal stride length and width. We minimize
\begin{align*}
 J(\mathcal{X})={}&\sum_{k=2}^{N+1}w_{z,k}
 \|\boldsymbol{z}_k-\boldsymbol{z}_{\mathrm{goal}}\|^2 \nonumber\\
 &+w_{\sigma}(\sigma_1-\sigma_{\mathrm{nom}})^2 \nonumber\\
 &+\sum_{k=1}^{N-1}\left[w_x(s_k^x-s_{\mathrm{nom}}^x)^2
 +w_y(s_k^y-s_{\mathrm{nom}}^y)^2\right]. 
\end{align*}
The nonnegative weights determine which target gives way when the terrain prevents all three from being met. Only the current timing is penalized because later gains are fixed at their nominal value by~\eqref{eq:sigma_fix}.

\titledparagraph{Capturability Bounds.}
Capturability concerns whether subsequent steps can bring the robot to rest. Computing exact capturable sets on a union of steppable regions is difficult, so we use linear bounds motivated by the DCM template. These constrain the predicted motion but do not certify that a capture sequence exists on the perceived terrain. They apply from $k=2$, since $\boldsymbol{z}_1$ and $\mathbf{p}_1$ are fixed inputs rather than decisions.

For lateral motion, the one-step capture condition without leg crossing motivates keeping the DCM on the inner side of the stance foot:
\begin{align} \label{eq:lateral_safety}
     \Delta_y \le (-1)^{k+\ell}(z_k^y-p_k^y) \le W_{\max}-\Delta_y, 
\end{align}
for $k=2,\dots,N.$ The sign convention is the same as in~\eqref{eq:slacks}; it points toward the opposite foot. 
The outward motion of the DCM is constrained by the stride width upper bound $W_{\max}$, with the fixed margin $\Delta_y>0$ keeps the DCM inside this lateral corridor.


For sagittal motion, assuming future steps can use the maximum length $L_{\max}$ and minimum duration $T_{\min}$ gives an infinite-step capturability bound~\cite{koolen2012capturabilitybased}. We tighten it by a fixed margin $\Delta_x>0$:
\begin{align} \label{eq:long_capt}
 |z_k^x-p_k^x|\le\frac{L_{\max}}{\sigma_{\min}-1}-\Delta_x,
 \quad k=2,\dots,N.
\end{align}
The maximum stride sets how far the support can move at each transition. Taking the minimum duration gives the least DCM growth before that move. Longer available strides or shorter durations therefore permit a larger offset under this template bound.
We also restrict forward growth between consecutive step-start states to limit the forward change between successive states:
\begin{align} \label{eq:zx_growth}
 z_{k+1}^x-z_k^x\le L_{\max},\quad k=1,\dots,N-1.
\end{align}

\titledparagraph{Swing Reachability.}
During replanning the swing foot is already in the air, so a revised plan must remain trackable before touchdown. Changing the landing position or the duration shifts the swing reference, and the foot trajectory task can only close that shift through its feedback loop: the later the change is made, the less time
the foot task has to absorb it, and the further the landing
falls from the new target. We therefore admit a revision only when this remaining error is smaller than the margin $m_j$ of the region that holds the commanded landing position $\mathbf{p}_2^{\mathrm{cmd}}$, so that the foot stays inside the selected region even though it lands slightly off target. This bounds how far the target may move,
\begin{align} \label{eq:swing_reach}
 \|\mathbf{p}_2-\mathbf{p}_2^{\mathrm{cmd}}\|_\infty \le \rho(\tilde t),
\end{align}
and, out of the same budget, how much the duration may change. The allowance $\rho$ is the region margin discounted by how much of the swing has already been carried out at elapsed time $\tilde t$, computed from the assumption of the foot task as a critically damped second-order response. Thus, nearly any target early in the swing within reach is still admissible, while near touchdown $\rho$ shrinks to a fraction of the margin $m_j$ and the planner can do little more than confirm the step it has committed to. This is a surrogate for swing feasibility rather than a full swing-dynamics constraint.


\titledparagraph{Steppable Region Constraints.}
Region boundaries require clearance, and poorly observed terrain calls for a larger margin. The observation model~\eqref{eq:variance_model} assigns higher variance to grazing-angle measurements, while unobserved cells lose confidence during map updates. Using the geometry and height uncertainty supplied by~\secref{sec:perception}, we set
\begin{align} \label{eq:region_margin}
 m_j=\min\!\big(m_0+\kappa\bar\sigma_j,\; m_{\max}\big).
\end{align}
Here $\bar\sigma_j$ is the mean height standard deviation, over the cells that lie inside region $j$ and pass the steppability threshold, of a slope-aware height standard deviation that combines the cell's posterior standard deviation with the height error that one cell of misregistration would produce on the local gradient. The fixed term $m_0$ is a geometric buffer on the foot \emph{center}, which is the quantity $\mathbf{p}_k$ constrains; it does not cover the foot's own half-width or half-length. The gain $\kappa$ is the one-sided $95\%$ Gaussian quantile, so the second term is a chance constraint on the region's height converted into horizontal clearance, and the cap $m_{\max}$ keeps a poorly observed region from shrinking away on uncertainty alone. In the reported experiments $m_0=0.02$~m, $\kappa=1.645$, and $m_{\max}=0.06$~m. 
Binary variables assign each foot position to one perceived region:
\begin{align} \label{eq:region_assignment}
 \delta_{kj}\in\{0,1\},\quad \sum_{j=1}^{M}\delta_{kj}=1,
 \quad k=1,\dots,N.
\end{align}
Membership in the selected region is imposed by
\begin{align} \label{eq:bigM_regions}
 \mathbf{A}_j\mathbf{p}_k\le\mathbf{b}_j-m_j\mathbf{1}
 +M_{\mathrm{big}}(1-\delta_{kj}),
\end{align}
for $j=1,\dots,M$ and $k=1,\dots,N$. When $\delta_{kj}=1$, the inequality places the foot inside the shrunken region; otherwise the big-$M$ term relaxes it. Unit row normals in $\mathbf{A}_j$ make $m_j$ an inward displacement in metres. Greater uncertainty therefore leaves less admissible area, and a region that shrinks to the empty set cannot be selected. We choose $M_{\mathrm{big}}$ from the local workspace bounds to relax unselected regions without an unnecessarily loose bound. The planner minimizes $J(\mathcal{X})$ over $\mathcal{X}$ and $\{\delta_{kj}\}$ subject to the constraints above. The cost is quadratic and every constraint is affine, so fixing the region assignments leaves a convex QP; the binary assignments make the problem an MIQP.

\begin{figure*}[t]
    \vspace{2mm}
    \centering
    \begin{minipage}[c]{0.4\textwidth}
        \includegraphics[width=\linewidth]{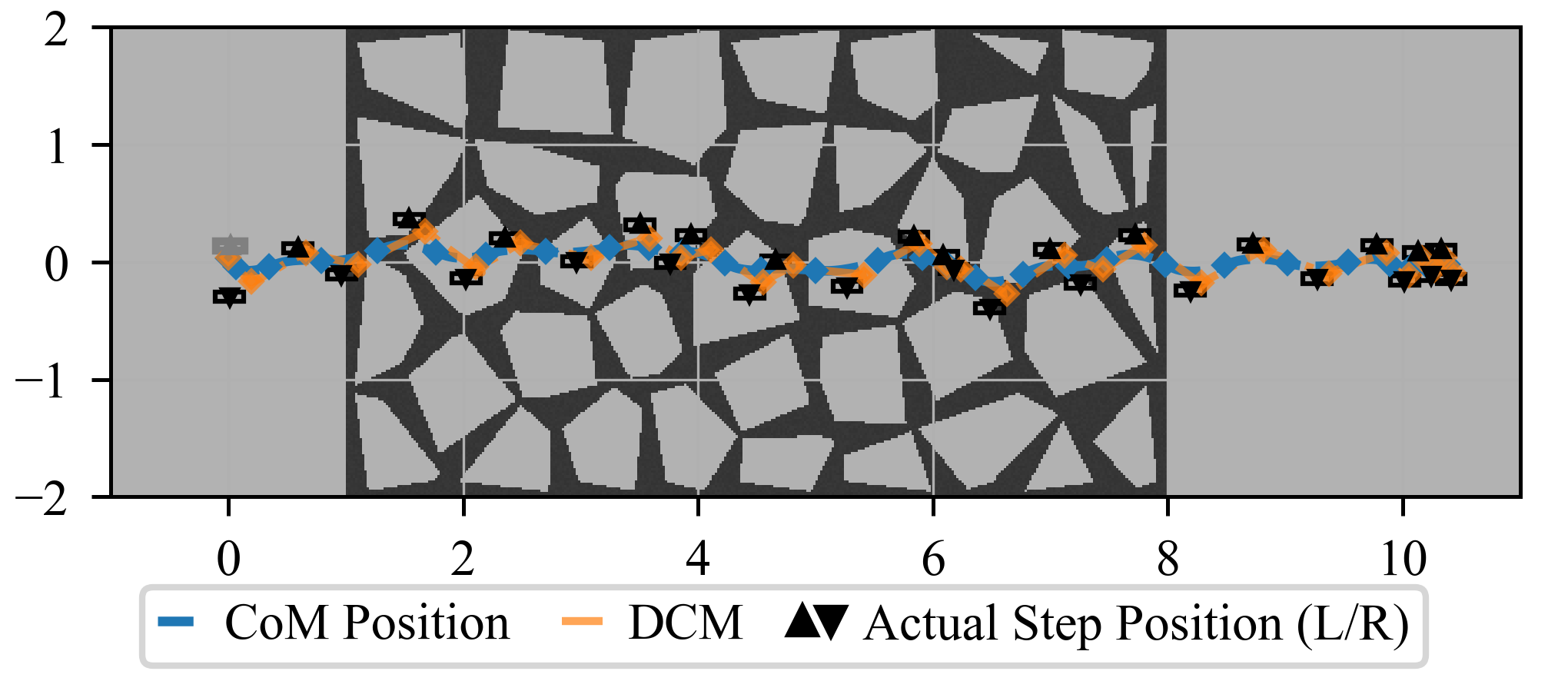}
    \end{minipage}\hfill
    \begin{minipage}[c]{0.28\textwidth}
        \includegraphics[width=\linewidth]{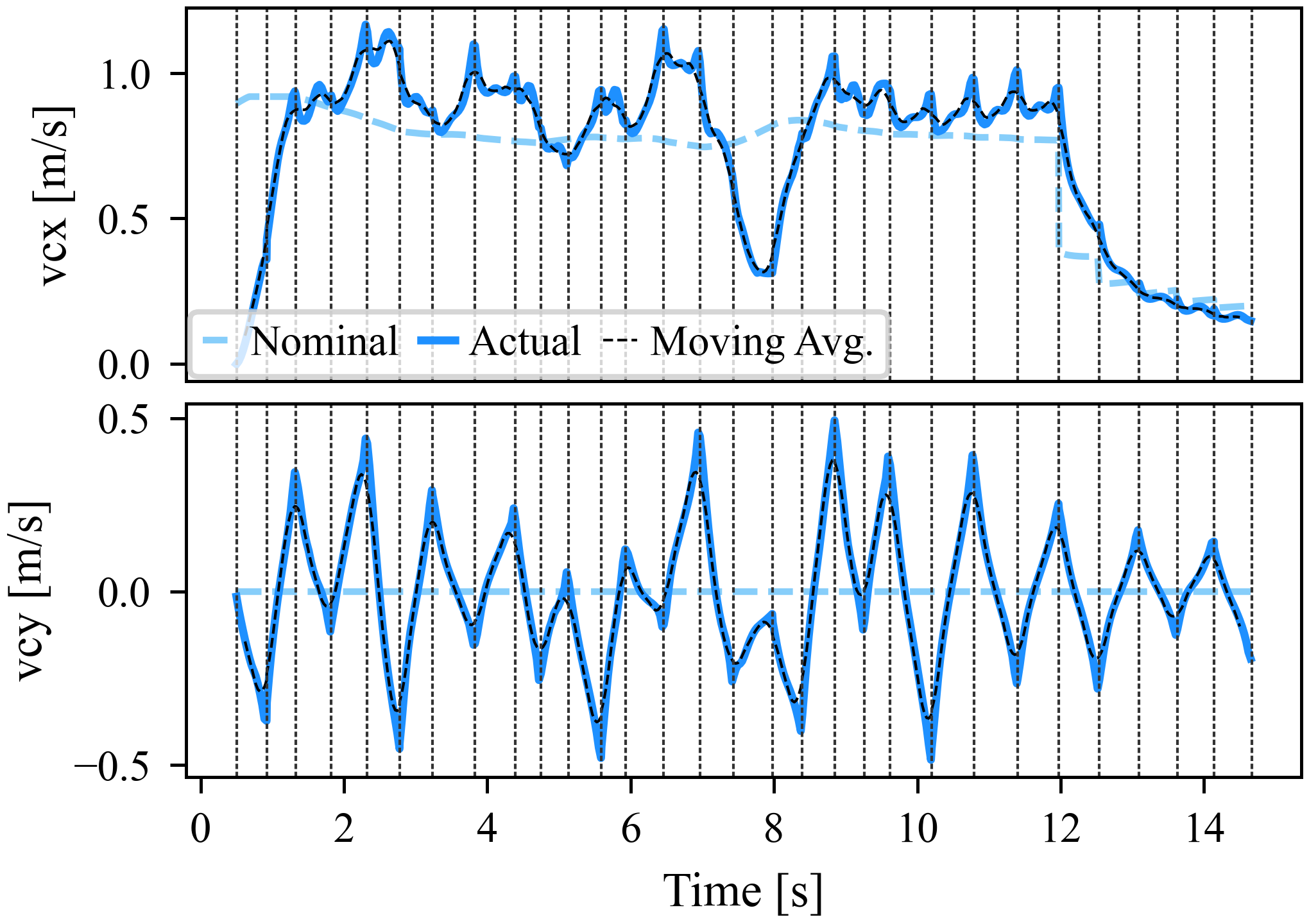}
    \end{minipage}\hfill
    \begin{minipage}[c]{0.28\textwidth}
        \includegraphics[width=\linewidth]{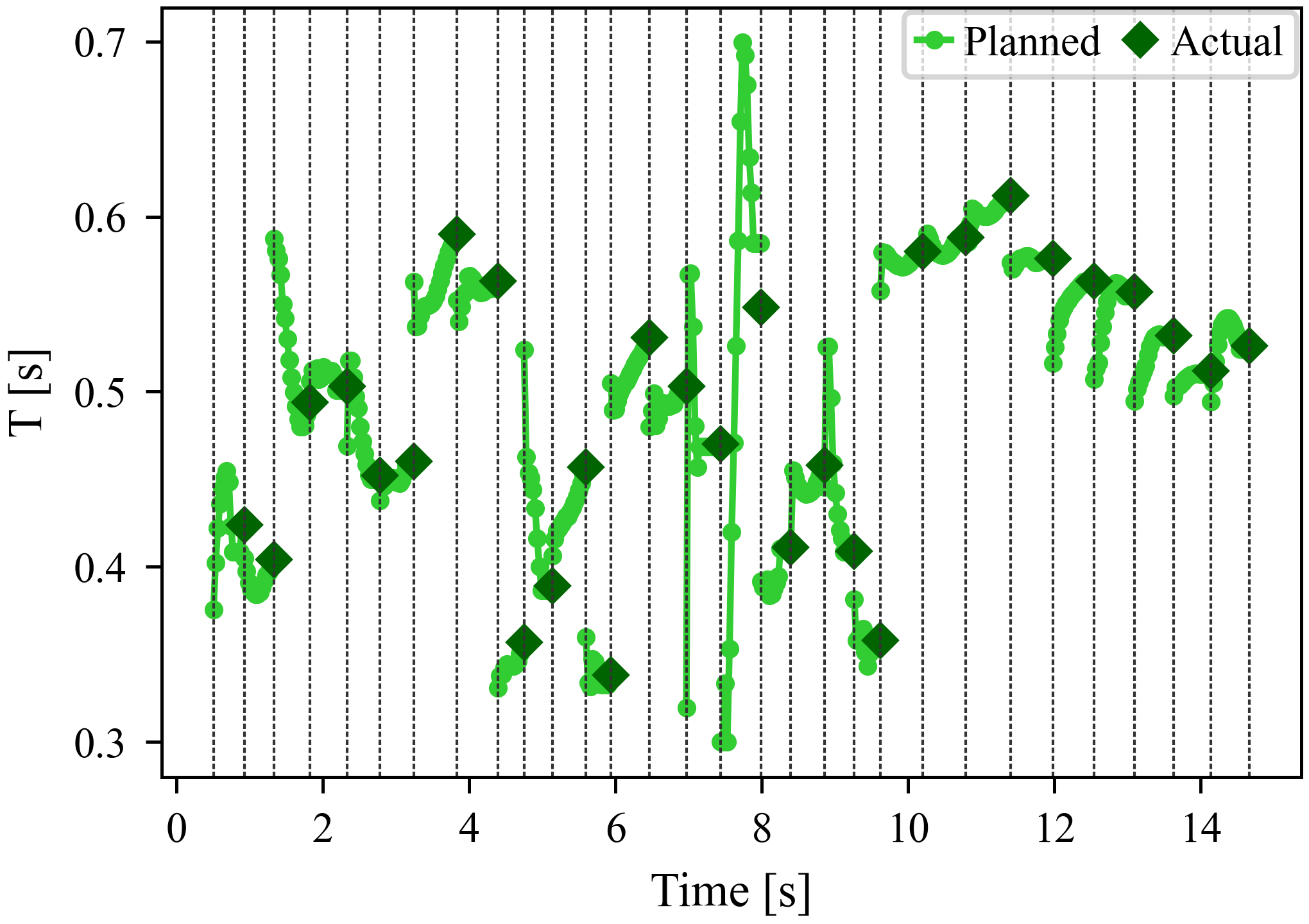}
    \end{minipage}
    \caption{A successful run of the proposed MIQP planner on one of the 100 randomized test fields. (Left) The robot walks from the left platform to the right platform through the stepping-stone region. The panel shows each landing foot position and how the CoM and DCM evolve. (Middle) Forward CoM velocity stays near 1~m/s over most of the crossing, although the planner does not track the hand-selected nominal velocity. It briefly dropped to about 0.3~m/s near 8~s, when the planner steps almost in-place before a large forward gap. (Right) The MIQP is solved again throughout each swing, and each solve adjusts the current step duration together with the landing position. The planned duration therefore changes throughout the swing, by up to about 0.2~s within one step. The actual duration (diamond) is close to the last planned value. Dotted lines mark step transitions.}
    \label{fig:sim_A_results}
    \vspace{-2mm}
\end{figure*}

\subsection{Within-Step Replanning and Execution}
Although the prediction advances between step starts, we update the plan during the swing using the latest measured DCM. Within a step, the template gives $\boldsymbol{\xi}(t)=\mathbf{p}_1+(\boldsymbol{z}_1-\mathbf{p}_1)e^{\lambda t}$. In the stance foot frame, $\mathbf{p}_1=\mathbf{0}$, so backward propagation from elapsed time $\tilde t$ gives the fixed initial state for the next solve:
\begin{align} \label{eq:z1_from_meas}
 \boldsymbol{z}_1=\boldsymbol{\xi}_{\mathrm{mea}}(\tilde t)e^{-\lambda\tilde t}.
\end{align}
This reconstruction makes the predicted DCM at the current time agree with the measurement under the template. Each solve therefore produces a different $\boldsymbol{z}_1$. It anchors the prediction to the present measurement rather than estimating the true state at the step start. If a disturbance changes the measured motion, the reconstruction changes accordingly, so the next solve starts from a prediction consistent with that motion.
Each solve uses this reconstructed DCM, the measured swing-foot position, and the latest available region data. Only the next landing position $\mathbf{p}_2$ and duration $T_1=\lambda^{-1}\ln\sigma_1$ are sent to the task-space whole-body controller~\cite{castillo2023template}; the later footsteps provide preview and are reconsidered in subsequent solves. Here $T_1$ is the total duration of the current step, so the intended time remaining at elapsed time $\tilde t$ is $T_1-\tilde t$.

The controller tracks constant base height and torso orientation, the swing-foot position reference, and zero swing-foot orientation in the stance frame. When the landing position or duration changes, the remaining minimum-jerk swing trajectory is recomputed to reach the revised target at touchdown. Its phase increases monotonically from $0$ to $1$, with the per-tick increment set by the current target duration and the phase derivative used to compute velocity commands. This allows the same trajectory interface to accept fixed or updated step durations.

\section{Results} \label{sec:results}
We evaluate how the framework combines foothold selection and motion planning on disconnected terrain. Simulation tests traversal of randomized stepping-stone fields and examines the roles of step timing, preview, and capturability bounds, in ablation with two proposed features. Hardware experiments then test the complete perception and planning pipeline on separated plates with two different arrangements. \footnote{A supplementary video of the results can be found at \url{https://xtificant.github.io/digit_fpp_pub/}.}

\subsection{Simulation Experiments} \label{sec:sim_results}

\titledparagraph{Setup.}
We use the Digit model in MuJoCo. The MIQP planner runs every 30~ms using Gurobi through its Python interface. Perception runs every 60~ms, which builds a probabilistic local heightmap using depth images from simulated renderers, and extracts the convex region constraints from it. The local heightmap covers $3\times3$~m with grid resolution $\Delta r=2$~cm. At most $M_{\max}=8$ regions, each with at most six edges, are passed to the planner. The planner uses an $N=4$ step preview, $T_{\mathrm{nom}}=0.5$~s, and $[T_{\min},T_{\max}]=[0.3,0.7]$~s. Stride bounds $[L_{\min},L_{\max}]=[-0.6,0.6]$~m and $[W_{\min},W_{\max}]=[0.05,0.5]$~m reflect the robot's kinematic limits. The DCM margins $\Delta_x=\Delta_y=0.02$~m were chosen using the foot size. Cost weights were tuned empirically to balance progress, timing, and stride objectives.


The test terrain fields is generating using a random circle packing method, which first pack circles of radius in $[0.25, 0.4]$ meters and spacing in $[0.02, 0.04]$ meters, in a rectangular area of $10\times4$ meters, and then inscribes polygons in each circle with numbers of vertices in $[4, 6]$ and random central angles. We generated 100 random test fields with different RNG seeds, and the left pane of~\figref{fig:sim_snapshot} shows one example of them with highly irregular gaps and sparse footholds. 
We compare the proposed MIQP planner (A) with two variants: fixed step duration (B) and reduced preview from $N=4$ to $N=2$ (C).



\titledparagraph{Walking Speed and Foot Placement.}
\figref{fig:sim_A_results} shows the results of a successful run of our framework on one of the test fields. The footprints in the left panel show the selected route, while the CoM and DCM trajectories show the accompanying body motion. Forward CoM velocity in the middle panel is around $1.0\,\mathrm{m/s}$ over much of the crossing, which is the result of chasing a fixed goal ahead with the nominal stride length and step duration objectives when the there are still smooth and feasible stepping plans through the region. However, at around 8 s, the forward velocity decreased to 0.3 m/s and the step duration rised to nearly the upper bound 0.7~s. As shown in the left panel, due to an significantly large gap in the forward walking direction, the planner yielded a solution to step in-place to keep balance and gain velocity for crossing this gap. The supplemental video shows the traversal.
In~\figref{fig:sim_A_results} (right), the planned duration changes both between steps and within swings. It reaches the lower bound when the robot steps back before the large gap, which helps keep balance on constrained footholds.
\begin{figure*}[t]
    \vspace{2mm}
    \centering
    \includegraphics[width=\textwidth]{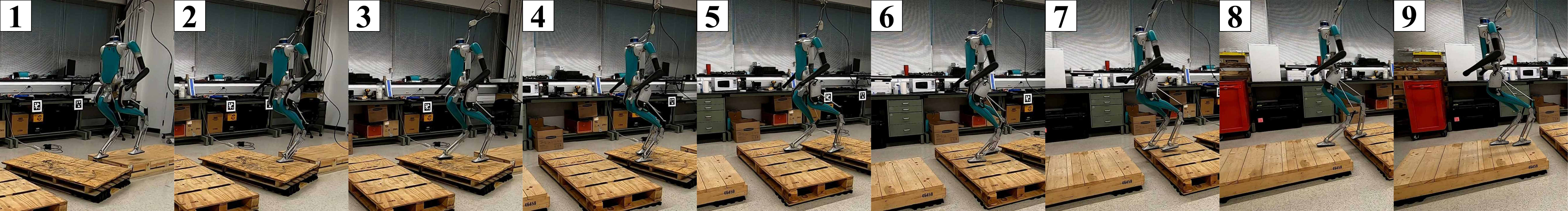}
    \caption{Snapshots in time sequence of the Digit walking through the test field, with the second plate rotated by $+15^\circ$. Key frames 1-3 show the moments when Digit was: starting a step across the first gap, landing the swing foot during the crossing, and walking inside the second plate. Key frames 4-6 and 7-9 show the moments near the second and third gap likewise.}
    \label{fig:hw_snapshots}
    \vspace{-2mm}
\end{figure*}

\begin{figure*}[t]
    \centering
    \includegraphics[width=\textwidth]{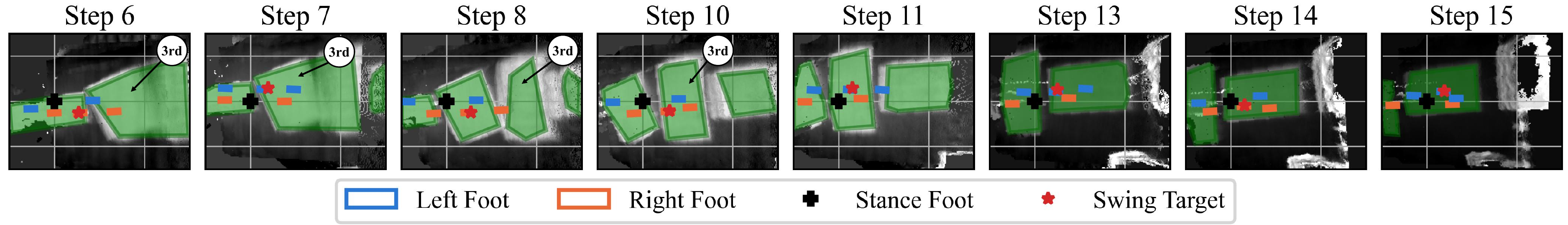}
    \caption{Hardware heightmaps at selected steps in the current stance foot frame. Lighter gray indicates higher terrain, so the white regions are the detected steppable plates; green polygons mark extracted region constraints; and the dashed blue rectangle is the selection window. The black cross marks the stance foot; blue/orange outlines denote a few left/right feet stance positions before and after the current stance position; and the red star marks the swing target.}
    \label{fig:hw_heightmaps}
    \vspace{-2mm}
\end{figure*}



\begin{table}[b]
    \caption{Ablation of the proposed foot-placement planner in MuJoCo}
    \label{tab:sim_ablation}
    \centering
    \begin{tabular}{ccccc}
        \toprule
        & \textbf{Crossing} & \textbf{Steps} & \textbf{Landing} & \textbf{Infeasible} \\
        \textbf{Configuration} & \textbf{success} & \textbf{to failure} & \textbf{success} & \textbf{solves} \\
        & (\%) & & (\%) & (\%) \\
        \midrule
        A & \textbf{83} & 24.0 & \textbf{99.3} & \textbf{5.8} \\
        B & 0  & 3.0 & -- & 79.6 \\
        C & 74 & 24.3 & 99.2 & 7.3 \\
        \bottomrule
    \end{tabular}
\end{table}

\titledparagraph{Ablation Results.}
The results of the 100 sweep tests of all three configurations are shown in~\tabref{tab:sim_ablation}. All three configurations are tested on the same 100 randomized terrains, and a test succeeds only when the robot reaches the far platform without falling. Even on such a challenging design of the test terrain fields and a harsh standard of finishing each test, the proposed framework (A) achieves a success rate of $83\%$. The second variant (B) fails to finish any of the tests, which usually falls right after stepping into the test field, and thus its success rate of landing inside regions is not worth reporting. There, it demonstrates that a variable step duration can significantly increase the range of viable evolutions of the DCM even on heavily constrained terrains, where the method with a fixed step duration could hardly plan a feasible stepping sequence. Compared to the third variant (C), longer preview horizons in our framework can better guide the DCM evolution within the viable range, resulting in a smaller rate of infeasible solve. This is because shorter preview horizons cannot obtain sufficient knowledge of the terrain further ahead, especially on such unstructured and sparse terrains, and thus brings difficulties in planning feasible paths.





\titledparagraph{Computation.}
The mean MIQP solve time in simulation for the proposed planner is 11.4~ms, with a 90th percentile of 16.8~ms. Both are below the 30~ms planner period, although these statistics do not establish that every solve meets that period. Branch-and-bound searches over discrete region assignments, producing variable runtimes. The eight-region limit controls problem size without imposing an upper bound on solve time. 


\subsection{Hardware Validation} \label{sec:hw_results}

\titledparagraph{Experimental Setup and Traversal.}
The hardware experiments use Digit with computation on a laptop equipped with an Intel i7 processor. Four body-mounted Intel RealSense D430/D435 cameras stream raw point clouds at 20~Hz. The planner and perception settings largely follow the simulation; the local map is enlarged from $3\times3$ to $4\times3\,\mathrm{m}$.

The platform consists of four rectangular wooden plates, with an average gap of $0.3\,\mathrm{m}$ and a uniform height of $0.18\,\mathrm{m}$ from the ground. In two experiments, the second plate is rotated by $+15^\circ$ and $-15^\circ$, respectively. This changes the edges and gaps encountered by the left and right swing feet. The robot crosses both arrangements, demonstrating execution with real depth data and altered foothold geometry. Snapshots of the hardware traversing the second arrangement are shown in~\figref{fig:hw_snapshots}.


\titledparagraph{Perception and Foot Placement.}
\figref{fig:hw_heightmaps} shows the local heightmap, extracted regions, and foot placements at selected steps. Each panel is expressed in its current stance foot frame. As the robot advances, the plates therefore move through the local map and their outlines change with the available depth observations. The green polygons represent the regions supplied to the planner, and the red star marks the current swing target. In every step, the near edges were always detected with higher confidence and accuracy than further ones, where nearby plates appear as separate polygons and the gaps are excluded. Although further gaps could be mistaken as steppable cells that are mixed with plates due to a larger distance and limited perspective, they will be correctly recognized and excluded as the robot approaches them, thus did not compromise the planning at any time. This can be observed by comparing the detection of the 3rd plate between in step 6/7/8 and step 10. 
Most steps land inside a plate, while a few land close to an edge. Likely contributors are errors in the estimated foot poses and in the relative transform between consecutive stance frames. Such errors misalign the map and the target with the physical plate. Foot tracking error may also shift the landing. Using the stance foot frame avoids a world-frame pose requirement, but still relies on estimates of relative motion.


\titledparagraph{Computation.}
During the two reported hardware experiments, the planner has a mean solve time of 4.0~ms and a 90th percentile of 5.1~ms over 685 solves. Both are well below the approximately 30~ms period of the 33~Hz planner. These times support real-time replanning on a laptop CPU: the landing position and duration can be revised about 16 times during a nominal 0.5~s step. The planner is faster than in simulation, likely because the plates produce fewer candidate regions and polygon edges than the stepping-stone field. Perception has a mean processing time of 20.8~ms and a 90th percentile of 28.2~ms over 661 evaluations. It therefore typically finishes before the next frame of the 20~Hz cameras arrives, 50~ms later.

\section{Conclusion} \label{sec:conclusion}

This paper presented a perceptive MIQP planner for walking on disconnected footholds. It selects a steppable region, the foot placement, and the current step duration together, using step-to-step DCM dynamics with capturability-based bounds. The regions come from a probabilistic heightmap built in the stance foot frame, and their margins grow with height uncertainty. The planner is solved again throughout each swing from the measured DCM. In simulation, Digit crossed randomized stepping-stone fields. On hardware, it crossed separated plates with the second plate rotated by $\pm15^\circ$, with a mean solve time of 4.7~ms.
The success rate on the hardest simulated fields is still limited, and a few hardware steps landed close to a plate edge, likely because of errors in the estimated motion between stance frames. Future work will add turning, handle terrain of variable height, and combine the planner with learning-based approaches.

\section*{Acknowledgement}
Generative AI tools were used in this work. ChatGPT (OpenAI) and Claude (Anthropic) assisted in code development and in revising the text in the paper.  All AI-assisted content was reviewed, edited, and verified by the authors.


\vspace{1em}

\bibliographystyle{IEEEtran}
\bibliography{references}

\end{document}